\documentclass[final, runningheads]{llncs}
\usepackage[T1]{fontenc}
\usepackage{multirow}
\usepackage{booktabs}
\usepackage{graphicx}
\usepackage{comment}
\usepackage{examples}
\usepackage{url}
\usepackage{tabularx}
\usepackage{amsmath}
\newcommand{\itadata}{\footnotesize \textsl{ITADATA2024: The 3$^{\text{rd}}$ Italian Conference on Big Data and Data Science}}
\usepackage{fancyhdr}
\fancypagestyle{empty}{%
  \fancyhf{}% clear
  \fancyhead[C]{\itadata}
  \fancyhead[R]{\includegraphics[width=.05\textwidth]{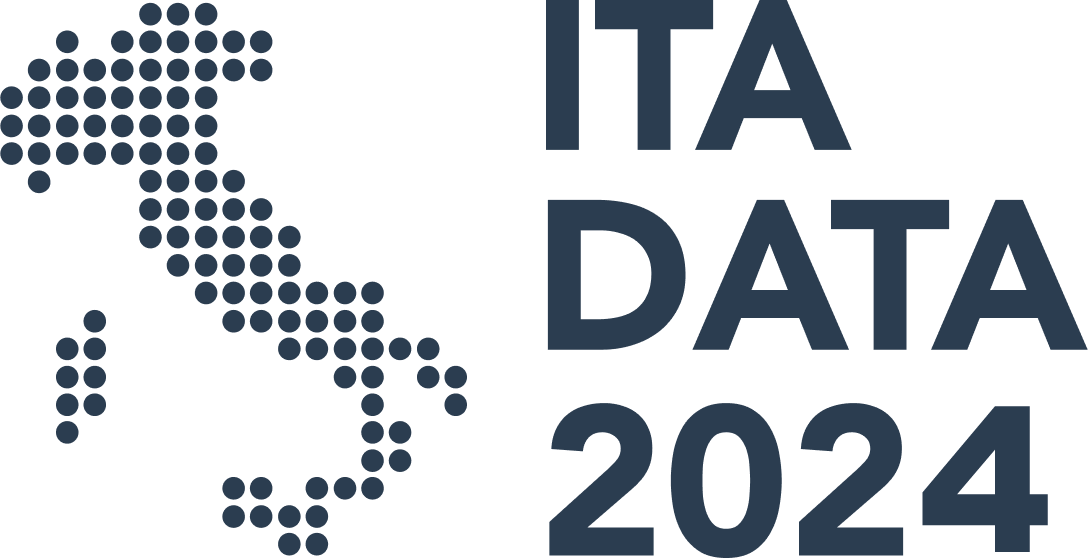}}% Your logo/image
}

\begin{document}
\title{From User Preferences to Optimization Constraints Using Large Language Models}
\author{Manuela Sanguinetti\orcidID{0000-0002-0147-2208} \\ \and
Alessandra Perniciano\orcidID{0009-0003-8956-5058} \and Luca Zedda\orcidID{0009-0001-8488-1612} \and Andrea Loddo\orcidID{0000-0002-6571-3816}  \and Cecilia Di Ruberto\orcidID{0000-0003-4641-0307} \and
Maurizio Atzori\orcidID{0000-0001-6112-7310}
}
\authorrunning{M. Sanguinetti and M. Atzori}
\institute{University of Cagliari, Department of Mathematics and Computer Science, Cagliari, Italy\\
\email{\{manuela.sanguinetti;alessandra.pernician;luca.zedda;andrea.loddo;dirubert;atzori\}@unica.it}}
\maketitle              % typeset the header of the contribution
\begin{abstract}
This work explores using Large Language Models (LLMs) to translate user preferences into energy optimization constraints for home appliances. We describe a task where natural language user utterances are converted into formal constraints for smart appliances, within the broader context of a renewable energy community (REC) and in the Italian scenario.
We evaluate the effectiveness of various LLMs currently available for Italian in translating these preferences resorting to classical zero-shot, one-shot, and few-shot learning settings, using a pilot dataset of Italian user requests paired with corresponding formal constraint representation. Our contributions include establishing a baseline performance for this task, publicly releasing the dataset and code for further research, and providing insights on observed best practices and limitations of LLMs in this particular domain.

\end{abstract}
\section{Introduction and Motivations}
The use of conversational agents in home automation and energy monitoring can be pivotal in household management, %enabling more efficient and resource-conscious habits. 
by making it easier for users to understand and control their energy consumption. The usefulness of such tools becomes even more evident in emerging contexts such as Renewable Energy Communities (RECs), in which members are also \textit{prosumers}, i.e., both consumers and producers of energy (typically from photovoltaic (PV) systems, sometimes also associated with storage batteries), and can share their excess energy flexibly with other members of the same community, thereby reducing the load from the main grid and overall energy costs. In such a scenario, the functions of a natural language-based interface can thus range from monitoring consumption and production trends, both at the level of the individual prosumer and at the level of the entire community, to generating advice, e.g., on the most appropriate times to use certain appliances in such a way as to maximize self-consumption (i.e., the use of energy from alternative sources than the main grid). The latter in particular can provide effective feedback when linked to an optimizer that can maximize self-consumption and thus enhance community-level efficiency. 

The work presented in this paper is part of a broader research project that aims to create an energy management system capable of managing large volumes of data related to both energy consumption and generation within a REC. In this context, the usefulness of a conversational interface lies in the possibility given to the user to consult this complex amount of data in an accessible and immediate way, taking advantage of the potential provided by dialogue systems (such as chatbots and virtual assistants). There are numerous challenges involved in the development of such a tool; these include the need to generate natural language messages from structured numerical data (relating to trends in energy consumption or production) \cite{Conde-Clemente-etal-2018}, or even the possibility of exploiting such data to create user profiles and tailoring the content to the appropriate profile, in order to deliver more effective customized guidance \cite{Trivino-Sanchez-2015}. %Recent advances in Natural Language Processing and in particular Large Language Models provide essential tools to address such challenges.
%Furthermore, 
Concerning in particular the optimization task, it is worth pointing out that user engagement and willingness to follow such guidance can also be influenced by how closely the agent's recommendations actually reflect the individual user's habits and preferences. Recent user studies have revealed how the optimization needs of a REC might poorly align with the actual needs of individual community members, especially when it comes to using certain appliances \cite{jensen-etal-2021,Pena-Jensen-2023_ecofeedback-RECs}. Therefore, it is essential to incorporate these preferences directly into the optimization function. This allows the optimizer to take such factors into account by using them as actual constraints, be they related to indoor comfort or time schedules related to users' individual routines. The ability of the conversational agent to properly encode such kind of information, and respond accordingly, is therefore essential for enhancing user’s compliance with more resource-aware practices. 

The integration of user preferences into energy management systems and appliance scheduling processes has been considered in previous studies within the field of operational research. This has been done either through the direct collection of preferences via static surveys \cite{Jin-etal-2017} or by considering user actions as a form of feedback with respect to suggested proposals \cite{Shuvo-Yilmaz-2022}. Little to no attention has been drawn so far to the extraction of constraints dynamically expressed by users through natural language. In fact, other frameworks have proposed the use of Natural Language Processing (NLP) techniques to extract relevant information from unstructured text regarding optimization problems, though pertaining to different topics \cite{Ramamonjison-etal-2023} (see Section \ref{sec:rel_work}). Our objective is inspired by the latter approach and it specifically aims to tackle a similar task within the broader project scenario mentioned above, and specifically within the Italian landscape. In particular, the main task we aim to address in our study is that of converting natural language usage preferences, expressed directly by users through the interaction with a conversational agent, into energy optimization constraints for household appliances. To address this task, we explore the potential offered by Large Language Models (LLMs) and in-context learning using a handcrafted pilot dataset of domain-specific user utterances and formal constraint representation. We prompted several LLMs among those available for Italian using classical zero-shot, one-shot and few-shot settings in order to assess their usefulness in adequately handling the mentioned task. The contribution we aim to propose is thus to provide an initial baseline on this task, also making available the data and code used for the experiments, along with some insights on recommended practices and approaches for effectively using LLMs in similar contexts, including the limitations we observed in our experiments.

The remainder of this the paper thus offers an introductory overview of the whole project along with a definition of the problem we intend to address in this work and the relevant body of literature. It will then provide a description of the data and the models used in our experiments and a discussion of the results obtained.

\section{Background and Problem Definition}

As mentioned above, this work is part of a larger project that involves, among other tasks, the development of a conversational tool to deliver energy feedback to members of a REC in Italy. The conversational interface has been designed to dynamically manage the various user requests, which can be related either to their own consumption trends or to the availability of alternative energy from their photovoltaic system or (if present) a battery storage system. Among the functions provided for this agent is the one related precisely to requesting suggestions on the best way to exploit one's energy resources, so as to maximize self-consumption. For this purpose the agent will then be connected to an external optimizer that aims to maximize the amount of shared energy within the REC as a whole. A prototype of this optimizer has been developed and described in Messilem et al. \cite{Messilem-etal-2024distributed} and Moré et al. \cite{More-etal-2024online}. This work aims to reconcile energy community-level goals with those of individual members, using the conversational interface to enable greater interaction between the user and the system. 

In light of the objectives and the usage scenario outlined above, we then defined the task at hand, i.e., given a natural language utterance expressing user's preferences or needs with respect to the use of a household appliance, the goal is to convert the user's expressed preferences into energy optimization constraints. The allowed constraints can be of two types: 1) \textbf{time} constraints, aiming at expressing that the user wants the appliance to be running at a certain time; 2) \textbf{temperature} setting constraints, that aim to further model the optimization module with preferences over the desired settings when the appliance of interest is a heat pump for both air conditioning and water heating. The current formulation assumes that the available input data comprises both the full user utterance (provided as context) and the %hte label of the appliance of interest,
text span (or spans) expressing the user preferences. %,% the label on the type of constraint (time or temperature) associated with the given text span. 
The intended output is thus a representation of the given preferences in the form of time or temperature constraints. %, expressed in the form of mathematical equalities and conditional statements. 
More formally, given a user utterance of $n$ words $w$ $U = \{w_1, ..., w_n\}$, and a set of text spans $C = \{c_1, ..., c_i\}$, where each span $c \subset U $ expresses a constraint in terms of usage preference, the objective is to map each constraint into a formal constraint representation $FR(c)$, thus having $FR(c) = map(c)$, $\forall c \in C$. The problem can thus be addressed as a classical sequence-to-sequence task, as it involves the conversion of a given natural language input into a structured format representing the constraints.

\section{Related Work}\label{sec:rel_work}
Our research approach builds upon recent work aiming at converting natural language descriptions into formal representations of optimization problems. The main rationale that lies behind such line of research consists in enabling a larger audience, possibly without any prior background on operational research, to successfully address optimization problems that may regard various aspects of everyday life. %The recent advances in Natural Language Processing (NLP) have provided a useful support, by making it possible to adapt large pre-trained language models to a wide range of different tasks, even with very few examples available. \\
%We have mainly drawn inspiration from the seminal
The primary contribution to this task can be found in the work by Ramamonijson et al. \cite{ramamonjison-etal-2022-augmenting}, who developed a benchmark dataset in English consisting of Linear Programming (LP) problems with textual descriptions of such problems. The dataset includes 1101 problems from domains such as advertising, investment, sales, manufacturing, science, and transportation. %The dataset is divided into training (713 examples), development (99 examples), and testing (289 examples). 
%The LP problems comprising the dataset are of the form:
%\begin{center}
%$
%\min_{x \in R^n} \mathbf{c}^\top \mathbf{x} \quad \text{s.t.} \quad \mathbf{a}_i^\top \mathbf{x} \leq b_i, \quad i = 1, \ldots, m
%$   
%\end{center}

%where $c$ and $a_i$ represent the parameters of the objective and the $i$-th constraint, respectively; $b_i$ is the right-hand-side limit, and the goal is to find $x$ that minimizes the objective value. 
In this framework, the objective and constraint functions are linear with respect to variables in the LP problems, and constraints are always expressed as inequalities. Three main representation formats have been designed to fully map, with a two-stage approach, the linguistic description into the actual mathematical formulation, thus going through an XML-like Intermediate Representation (IR) and a canonical (tabular) form. As a result, each example has a text description of the problem and is annotated with its corresponding IR, math representation, and canonical formulation. 

 The dataset was then used as benchmark within the NL4Opt shared task, that took place in 2022 at NeurIPS \cite{Ramamonjison-etal-2023}.\footnote{\url{https://github.com/nl4opt}. The second edition of the competition has recently been launched: \url{https://nl4opt.github.io/neurips-2024/}} The competition precisely consisted in translating textual descriptions of optimization problems into Linear Programming (LP) formulations. It was organized into two distinct sub-tasks: a sequence labeling task aimed at identifying all the entities in the text that were relevant for the optimization problem formulation (i.e., variables, objective function and constraints) and a sequence-to-sequence task aimed at generating the mathematical problem formulation. Sub-task 2 in particular uses both the input linguistic description along with the text spans identified in the previous sub-task to generate the problem formulation. The task organizers developed a separate baseline for each sub-task, using XLM-RoBERTa-base for the sequence labeling step and BART with a copy mechanism for generating the LP formulation, obtaining a macro-averaged $F1=0.906$ in the former and a declaration-level mapping accuracy of $0.610$ in the latter. Once the competition closed, the organizers finally investigated how ChatGPT would perform on the same data. They bypassed the competition's subtasks and directly asked ChatGPT (model GPT-3.5-turbo) to generate the problem formulations from the given text descriptions, with human experts assessing the output quality of the model on the test set. The model reached a declaration-level accuracy of $0.927$.

The best-performing system in Subtask 2 \cite{Gangwar-Kani-2023} obtained an accuracy of $0.899$ with a “decode all-at-once” strategy; the textual input was enhanced surrounding variable entities with XML-tags, and both objective and constraint declarations were generated simultaneously using BART-large. 
%Zong and Krishnamachari (2023) evaluated the performance of Generative Pre-trained Transformer 3 (GPT-3, Brown et al., 2020) in three types of tests: classifying mathematical word problems that can be expressed as two linear equations; extracting equations from word problems; and generating word problems. The accuracy of extracting equations increases with the number of examples provided in the prompt, and the accuracy of the fine-tuned GPT-3 is higher than that of the few-shot learning and zero-shot learning. GPT-3 with zero-shot learning performs well on classifying word problems for most classes, and their experiment results provide some evidence that supplying examples of the same class in the prompt for few-shot learning for extracting equations can be helpful. However, in contrast to our approach, they do not incorporate classification as a prior step to equation extraction in their framework. Notice that problem classification is not always helpful in equation extraction. Compared to our approach, these authors fine-tuned GPT-3 used for equation extraction rather than for the classification task with a limited number of valid responses. (FROM LI ET AL 2023)

Li et al. \cite{Li-etal-2023} further extended the NL4Opt approach %by decomposing the development data released for the competition to clearly separate objectives and constraints, and 
by expanding the dataset with brand new descriptions that would include logic constraints (e.g., $if~A~then~B$) and equality constraints. %This resulted in a dataset of 574 pairs of problem descriptions and corresponding variable definitions, that they used for training and validation.%, divided into 463 for training and 101 for validation. 
%An additional brand new test set was finally created from scratch. 
To address the task, the authors adopted a three-stage method that employs ChatGPT for variable identification, a fine-tuned GPT-3 model for constraint classification, and again ChatGPT for constraint generation. For stages 1 and 3 ChatGPT was prompted with a zero-shot setting. Alternatively, PaLM 2 was also fine-tuned for all stages, though obtaining lower results overall.

Although our work shares the same basic goal with those just described, the specifics of our use case mean that our approach diverges significantly with respect to several points: (1) the type of input expected is a single user utterance, not the description of an optimization problem; (2) the task definition, at least at this stage, involves the conversion of constraints only, whether they are related to temperature or to the desired operation times of the appliance; and (3) the representation of constraints, as described below, involves an abstract formulation that does not conform with the canonical form of linear constraints. 
 %However, both NL4Opt's authors and Li et al. \cite{Li-etal-2023} focus on inequality constraints within specific domains like advertising and manufacturing. Our work differs by tackling the translation of user preferences into energy-saving constraints for home appliances, requiring in particular specific formulation for time constraints.

\section{Data and Annotation Scheme}\label{sec:scheme}
The data sample used in these experiments comes from a larger domain-specific corpus of user utterances and labeled intents, that was primarily used to train the NLU module of the conversational agent introduced above.\footnote{The backbone of the agent has been implemented with RASA \cite{bocklisch-etal-2017-rasa}, and for the NLU module the built-in Dual Intent and Entity Transformer (DIET) \cite{bunk-etal-2020-diet} classifier was used.} The overall corpus consists of 157 utterances in Italian grouped into three intent categories, two for monitoring energy consumption and PV production, and one for requesting optimization advice from the agent.\footnote{The whole dataset in YAML format uset to train DIET is available here: https://github.com/msang/nl-interface/tree/main/data} Out of the former category, some user requests involve general questions on how best to use their appliances, as in Example \ref{ex:utter_1} below, while a pilot sub-sample (26 utterances) includes requests where specific preferences, in terms of time ranges or temperature settings, are expressed, as in Example \ref{ex:utter_2}:

\begin{examples}
    \item \textit{quando mi conviene aumentare l'acqua al massimo?} \\
    EN: "When is it convenient for me to set the water heater to the maximum?" \label{ex:utter_1}
    \item \textit{come risparmiare tenendo il climatizzatore sempre acceso?} \\
    EN: "How to save money by keeping the air conditioner always on?" \label{ex:utter_2}
\end{examples}

This pilot sub-sample of 26 utterances then underwent a two-step manual annotation to identify and label the text spans within each utterance that express such kind of preferences, and then to convert these preferences into possible constraint representations, that can be further processed and fed to the external optimization module.

The constraint annotation scheme considers the following parameters and variables, some of which are derived directly from the pre-existing definitions of the optimizer. In particular, we consider:

\begin{enumerate}
    \item Indices and variables: 
    \begin{itemize}
        \item  $T = \{1, ..., t\}$: discrete set of intervals that constitute the time horizon T
        \item  $s_t \in \{0,1\}$: binary variable indicating the state of the appliance at time t
        \item $h_t \in [H_{min}, H_{max}]$: continuous variable representing the desired temperature at time t, where $H_{min}$ and $H_{max}$ are the minimum and maximum temperatures allowed by the system
    \end{itemize}
    \item User-specified parameters:
    \begin{itemize}
        \item $user_{start}$, $user_{end}$: desired start and shutdown times of the appliance (either both or only one of them can be expressed by the user)
        \item $user_{temp}$: temperature desired by the user
    \end{itemize}    
\end{enumerate}

The intermediate constraint representation format involves first assigning the value of the decision variable (be it $s_t$ or $h_t$), and then the value of $t$ for which that variable takes the assigned value.

Based on the recurring patterns observed in the pilot set of utterances, the following conditions can be outlined:

\begin{itemize}
    \item The user wants the household appliance to run all the time: \\
    \vspace{0.2cm}
        $s_t = 1  \forall t$
    \item The user wants the household appliance to be on/off at precise time intervals:\\
    \vspace{0.2cm}
        $s_t = 1 \lor 0 \forall user_{start} \leq t \leq user_{end}$
    \item The user wants the appliance to be turned on/off starting from a specified time:\\
        \vspace{0.2cm}
       $s_t = 1 \lor 0 \forall t \geq user_{start}$
    \item The user wants the appliance to be turned on/off until a specified time:\\
        \vspace{0.2cm}
        $s_t = 1 \lor 0 \forall t \leq user_{end}$
\end{itemize}
The same scheme can be applied to the temperature variable, where the value assigned will be the one specified by the user, thus having $h_t=user_{temp}$.

\begin{figure}
    \centering
    \includegraphics[scale=0.5]{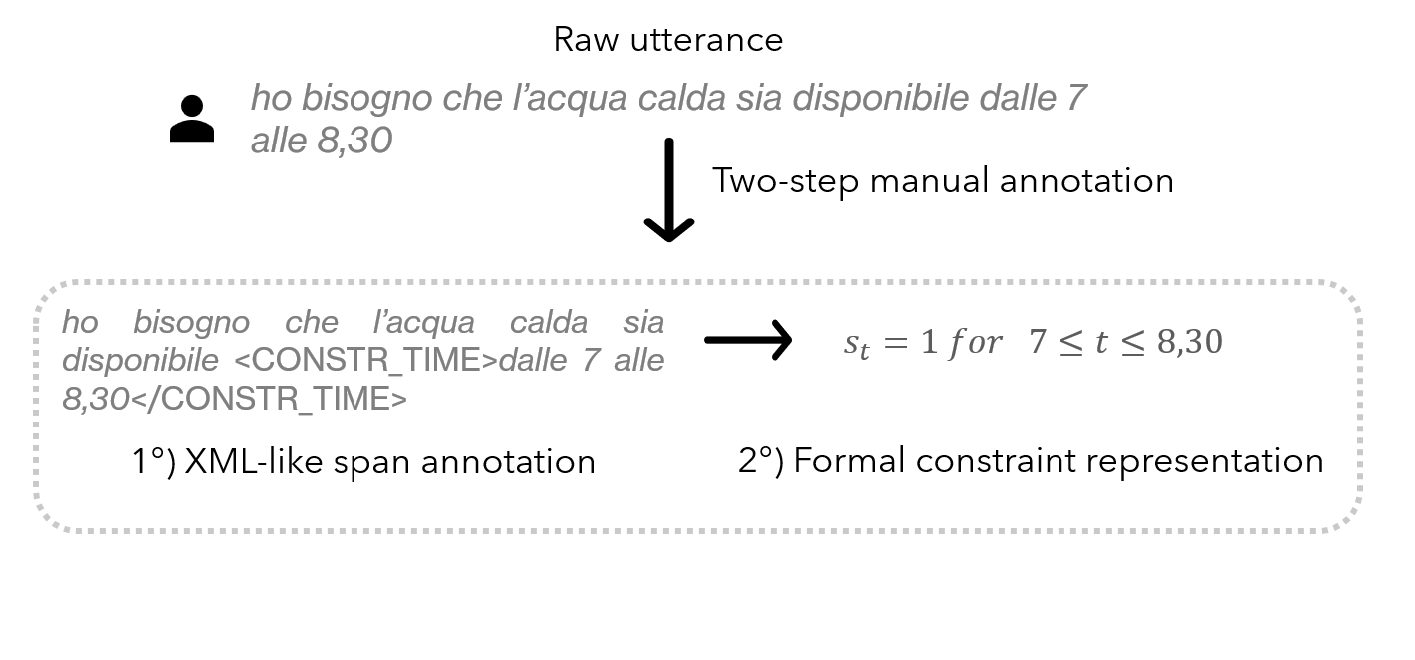}
    \caption{Annotation pipeline and example for the user utterance "\textit{ho bisogno che l’acqua calda sia disponibile dalle 7 alle 8,30}" (EN: "I need hot water to be available from 7:00 to 8:30.")}
    \label{fig:annot_ex}
\end{figure}

The two-step annotation pipeline along with an actual example from the dataset are shown in Figure \ref{fig:annot_ex}.

\begin{comment}
%The resulting annotation can be shown in Examples \ref{ex:annot_1}--\ref{ex:annot_2}.
\begin{examples}
    \item \textit{...}\\
    EN: "" \label{ex:annot_1}
    \item \textit{...} \\
    EN: "" \label{ex:annot_2}
\end{examples}
\end{comment}

In the next sections we will describe how we used the annotated data with a set of Large Language Models currently available for Italian.

\section{Models and Settings}
Given the wide variety of tasks and application domains for which they can be used, we decided to explore the capabilities of Large Language Models (LLMs) on this specific task as well. In particular, we chose to use more recent instruction-tuned models, which allow a description of the task to be provided within the prompt fed to the models. This approach was taken to optimize model performance in the absence of sufficient data for tackling standard fine-tuning approaches, as in our case. In addition, the scarcity of training data is compounded by the limited availability of usable models of this type for the Italian language. We thus preliminary selected for our experiments the following LLMs: ChatGPT (model GPT-3.5-turbo), LLaMAntino-2-UltraChat \cite{Basile-etal-2023-llamantino2}\footnote{Due to hardware constraints, the smaller version -- 7B parameters -- was used.} and LLaMAntino-3-ANITA \cite{Polignano-etal-2024-llamantino3} (fine-tuned versions of Llama2 and Llama3, respectively), and Zefiro\footnote{\url{https://huggingface.co/giux78/zefiro-7b-beta-ITA-v0.1}}, a fine-tuned model of Mistral \cite{Jiang-etal-2023-mistral7b}. For the former model we used the online interface,\footnote{\url{https://openai.com/chatgpt}} while for the others we used the model checkpoints available through the HuggingFace platform and using the Transformers library. 

\paragraph{Prompt structure and settings} Drawing inspiration from the approach followed in Gangwar and Kani \cite{Gangwar-Kani-2023} (see Section \ref{sec:rel_work}), the input utterances included in each prompt were labeled with XML tags that incorporate the text spans expressing users' preferences. In order to guide the models towards the appropriate  constraint generation, the prompts were structured so as to include: i) a brief introduction to the task; ii) the meaning of the XML tags; iii) the constraint representation format; iv) in the case of one and few-shot prompts, one or five pairs of XML-tagged utterances and corresponding constraint formal representation; v) the input utterance. 
As mentioned above, we queried each LLM with different prompts, with the aim of further enhancing the models' performance. We empolyed common one-shot and few-shot approaches, including additional examples (up to 5 in the few-shot setting) within the prompt. 

In order to limit randomness in the responses generated by the various models, while ensuring they select a diverse yet relevant set of tokens, for the open models (thus excluding ChatGPT) we used the following combination of decoding parameters: temperature = 0.1, top-k = 20, and top-p = 0.9. In addition, to keep the responses short and concise, we set the maximum generation length to 30 tokens.\footnote{Both annotated data in JSONL format, prompts and Python code used for the experiments are available at the following GitHub page: \url{https://github.com/msang/nl-interface/tree/main/nl2optim}}

\paragraph{Evaluation metrics} Due to the specific nature of this task, standard text-to-text generation metrics, such as BLEU, ROUGE or METEOR, were not deemed suitable. To measure the closeness between the output generated by the various models tested and the formal constraint representation described in the previous section, we adopted two specific metrics. One is ChrF \cite{Popovic-2015-chrF}, which considers the overlap of character n-grams between reference and generated constraints. The measure is computed as follows:

\begin{center}
 $\text{ChrF}_\beta = \frac{(1 + \beta^2) \cdot \text{ChrP} \cdot \text{ChrR}}{\text{ChrR} + \beta^2 \cdot \text{ChrP}}$   
\end{center} 

where $ChrP$ and $ChrR$ are character n-gram precision and recall averaged over all n-grams, and $\beta$ is a weight assigned to recall (with $\beta$ = 1, precision and recall have equal importance). Although the task at hand cannot be properly defined as text-to-code, the adherence to a format with a specific set of symbols and a precise syntax makes it somehow similar to the latter. Hence the choice to use a metric that could faithfully measure such adherence in the generated constraints. Recent studies also suggest that, namely in the context of text-to-code generation, this measure is preferable to other standard metrics \cite{Evtikhiev-etal-2023}.

In addition to character matching, we also aimed to evaluate the accuracy of the model in correctly identifying both the variables involved (i.e. $s_t$ for usage timing preferences and $h_t$ for temperature preferences) and the conditions expressing the time value $t$. To do this, we parsed the generated responses to extract these two aspects and calculated the accuracy of the models with respect to their correct generation. We thus computed accuracy separately for variables and conditions:

\begin{equation}
Acc_{Variables} = \frac{1}{N} \sum_{i=1}^N \frac{Variables^{gen}_i}{Variables_i}   
\end{equation}

\begin{equation}
Acc_{Conditions} = \frac{1}{N} \sum_{i=1}^N \frac{Conditions^{gen}_i}{Conditions_i}   
\end{equation}

where $N$ is the number of utterances used for evaluation, $Variables_i$ and $Conditions_i$ represent the total number of variables and conditions, respectively, in the constraint formulations for utterance $i$, and $Variables^{gen}_i$ and $Conditions^{gen}_i$ represent the number of variables and conditions correctly generated by the model for utterance $i$. 

Next section reports and discusses the results obtained.

\section{Results and Discussion}
As mentioned above, four models available for Italian were tested with the various prompts. However, of these, both LLaMAntino-2 and Zefiro returned inconsistent and noise-rich results. We therefore considered their outputs to be invalid for evaluation and were discarded. On the other hand, the outputs generated by ChatGPT and LLaMAntino-3 were evaluated using the metrics reported in the previous section. The results obtained are shown in Table \ref{tab:results}. 

\begin{table}[]
    \centering
    \begin{tabular}{l|c|>{\centering}p{0.15\textwidth}|c|c|c}
    \toprule
    & prompt & ChrF & Acc$_{Variables}$ & Acc$_{Conditions}$ & Acc$_{Avg}$ \\
    \midrule
    \multirow{3}{*}{Cerbero} & \textbf{0s} & 43.0734 & 0.7381 & 0.1190 & 0.4286 \\
                              & \textbf{1s} & 35.9382  & 0.2619 & 0  & 0.1310 \\
                              & \textbf{fs} & -- & -- & -- & -- \\
    \midrule
    \multirow{3}{*}{ ChatGPT } & \textbf{0s} & 51.0562 & 0.7857 & 0.2143 & 0.5 \\
                               & \textbf{1s} & 60.8065 & 0.7857 & 0.119 & 0.4524 \\
                               & \textbf{fs} & 69.0289 & 0.7619 & 0.3571 & 0.5595 \\
    \midrule
    \multirow{3}{*}{LLaMAntino-3-ANITA} & \textbf{0s} & 37.9288 & 0.5 & 0.0714 & 0.2857 \\
                              & \textbf{1s} & 66.238 & 0.7222 & 0.2857 & 0.504 \\
                              & \textbf{fs} & 74.5472 & 0.8571 & 0.4286 & 0.6429 \\
    \midrule
    \multirow{3}{*}{Maestrale} & \textbf{0s} & 33.8217 & 0.2063 & 0.0238 & 0.1151 \\
                              & \textbf{1s} & 59.0048  & 0.7619 & 0.3571  & 0.5595 \\
                              & \textbf{fs} & -- & -- & -- & -- \\
    \midrule
    IT5                & --          & 47.0815 & 0.4545 & 0.2 & 0.3273 \\
    \bottomrule
    \end{tabular} 
    \caption{Results of the tested models with zero-shot (0s), one-shot (1s) and few-shot (fs) prompts. }
    \label{tab:results}
\end{table}

These results show the overall inadequacy of the tested LLMs to address this task using only the in-context learning approach; given the particular nature of the task, further supervised fine-tuning is therefore required. On the other hand, going into more detail about the results with respect to individual metrics and prompt settings, it is possible to see how ChatGPT obtains better results than LLaMAntino-3 when only the task instructions and the input utterance are provided with the preferences embedded within the XML tags. However, when further examples are given, LLaMAntino-3 shows better performance in terms of both overlapping char-grams and average accuracy. As for the accuracy in generating both the variables and the conditions that determine the value of $t$, it can be observed that both models have more problems generating conditions correctly than variables. In the case of ChatGPT in particular, the addition of a single annotation example reduced the model's performance in generating conditions, while leaving those related to variable generation unchanged. Performance did not benefit from the additional context provided, suggesting that ChatGPT may need more examples or different prompt optimization strategies to improve significantly. In case of LLaMAntino-3, on the other hand, the model improves incrementally -- both in the generation of variables and conditions -- as the number of examples in the prompt increases, showing, at least in this setting, greater capabilities when provided with more context.

\section{Conclusions}
In this paper, we explored the capabilities of some LLMs available for Italian in translating user preferences into energy optimization constraints for household appliances. Given the current scarcity of data, in this phase of the work we preliminarily evaluated the effectiveness of these models using only prompt engineering techniques. We gradually increased the number of examples provided to each model, analyzing the differences in performance at each step. The results obtained clearly indicate that, for this specific task, a more in-depth knowledge of the specific domain is required. This involves the use of supervised fine-tuning techniques. As future work, we plan to further expand the dataset of utterances expressing user preferences and annotate the related constraints; we intend to do so replicating the two-step approach also adopted for developing the pilot dataset, as also described in Section \ref{sec:scheme}, thus first manually identifying the text spans expressing the user preferences and then converting them into formal constraint representations. Finally, concerning the evaluation part, a further extension could be applied by adopting, once the model is effectively trained and properly connected with the actual optimizer, measures to also evaluate the functional correctness of the constraints produced. This approach could take inspiration from metrics such as \textit{pass@k}, which harks back to the practices of human developers, who judge the correctness of code based on passing a series of unit tests. This in particular would serve to highlight the effectiveness of the task in relation to the specific application scenario of the overarching project.

\begin{credits}
\subsubsection{\ackname} This work has been developed within the framework of the project e.INS- Ecosystem of Innovation for Next Generation Sardinia (cod. ECS 00000038) funded by the Italian Ministry for Research and Education (MUR) under the National Recovery and Resilience Plan (NRRP) - MISSION 4 COMPONENT 2, "From research to business" INVESTMENT 1.5, "Creation and strengthening of Ecosystems of innovation" and construction of "Territorial R\&D Leaders". This work was also partially funded under the National Recovery and Resilience Plan (NRRP) -  Mission 4 Component 2 Investment 1.3, Project code PE0000021, “Network 4 Energy Sustainable Transition--NEST”.

%\subsubsection{\discintname}
%It is now necessary to declare any competing interests or to specifically state that the authors have no competing interests. Please place the statement with a bold run-in heading in small font size beneath the (optional) acknowledgments\footnote{If EquinOCS, our proceedings submission system, is used, then the disclaimer can be provided directly in the system.}, for example: The authors have no competing interests to declare that are relevant to the content of this article. Or: Author A has received research grants from Company W. Author B has received a speaker honorarium from Company X and owns stock in Company Y. Author C is a member of committee Z.
\end{credits}
 \bibliographystyle{splncs04}
 \bibliography{main}
%Author, F.: Article title. Journal \textbf{2}(5), 99--110 (2016)

%Author, F., Author, S.: Title of a proceedings paper. In: Editor,
%F., Editor, S. (eds.) CONFERENCE 2016, LNCS, vol. 9999, pp. 1--13.
%Springer, Heidelberg (2016). \doi{10.10007/1234567890}

%Author, F., Author, S., Author, T.: Book title. 2nd edn. Publisher,

%Author, A.-B.: Contribution title. In: 9th International Proceedings
%on Proceedings, pp. 1--2. Publisher, Location (2010)

%LNCS Homepage, \url{http://www.springer.com/lncs}, last accessed 2023/10/25
%\end{thebibliography}

\end{document}